\documentclass[11pt,letterpaper]{article}
\usepackage[margin=0.8in]{geometry}
\usepackage[T1]{fontenc}
\usepackage{lmodern}
\usepackage{cite}
\usepackage{amsmath,amssymb,amsfonts}
\usepackage{algorithm}
\usepackage{algorithmic}
\usepackage{graphicx}
\usepackage{booktabs}
\usepackage{array}
\usepackage{textcomp}
\usepackage{float}
\usepackage[section]{placeins}
\usepackage[hidelinks]{hyperref}
\makeatletter
\g@addto@macro\UrlBreaks{\do\-}
\makeatother

\graphicspath{{figure/}}
\title{EPOC: Endpoint-Preserving Online Correction With Compressed Residual State for Multi-Horizon Time Series Forecasting}
\author{Takumi Fujimoto$^{1}$ \qquad Hiroaki Nishi$^{2}$\\[0.5em]
\small $^{1}$School of Science for Open and Environmental Systems,\\
\small Graduate School of Science and Technology, Keio University, Japan\\
\small \texttt{takmin@keio.jp}\\
\small $^{2}$Department of System Design, Faculty of Science and Technology,\\
\small Keio University, Japan\\
\small \texttt{west@keio.jp}}
\date{}
\hypersetup{
  pdftitle={EPOC: Endpoint-Preserving Online Correction With Compressed Residual State for Multi-Horizon Time Series Forecasting},
  pdfauthor={Takumi Fujimoto and Hiroaki Nishi}
}

\begin{document}
\maketitle

\begin{abstract}
Completed forecasts provide residual feedback, but retaining full residual blocks increases auxiliary state. We propose Endpoint-Preserving Online Correction (EPOC) with a compressed residual state. It stores low-order discrete cosine transform (DCT) coefficients and the final value of the preceding residual block. Each channel shares the endpoint across component-wise online ridge regressions using current-forecast coefficients. We evaluate eight multivariate series, DLinear and PatchTST, three seeds, and two training variants: 96 matched fixed-base conditions at a 24-step horizon. EPOC achieves mean condition-wise reductions in mean squared error (MSE) and mean absolute error (MAE) of 15.40\% and 9.35\% from the uncorrected base, respectively, with a median of 6,352 B in retained auxiliary arrays. EPOC also outperforms the $\delta$-Adapter, COSA, FAC, and OMPB in paired MSE on most conditions while retaining less state. Full ELF achieves the largest mean MSE reduction, 19.29\%, but its median retained state is 474,048 B ($\times75$ relative to EPOC). Equal-size summary controls favor the endpoint by 1.65--2.20\% in paired MSE; a coefficient-reconstructed endpoint yields similar accuracy to the observed endpoint, highlighting its role as a shared input. Increasing the retained DCT component count from 4 to 8 adds 1.00 percentage point of MSE reduction for 5,728 B. On jointly trained bases, EPOC lowers MSE by 16.69--20.15\% relative to globally blended TEFL-style adapters applied to the same base. Code and numerical records are available at \url{https://github.com/keiotakmin/endpoint-preserving-residual-correction}.

\end{abstract}


\section{Introduction}
\label{sec:introduction}

A multi-horizon forecaster issues predictions before their targets arrive. Once a block is complete, its errors can guide the next prediction. With a fixed base forecaster, an auxiliary correction must decide which past errors to retain between forecasts and how much state to devote to them.

Residual feedback has been used to correct traffic forecasts~\cite{rescal2022}, while TEFL~\cite{tefl2026} feeds a fully observed multi-step residual vector to a learned adapter trained jointly with the base model. Other fixed-forecaster approaches learn an auxiliary online forecast from Fourier features of observation history~\cite{elf2025}, correct a forecast using recent target statistics~\cite{cosa2026}, or calibrate input and output spectra~\cite{fac2026}. These methods use different information about the recent stream. We study a compressed representation of the base forecaster's own errors that keeps the most recent error explicitly available.

Keeping a complete residual block preserves every horizon-specific error but requires state proportional to its length. A few transform coefficients require less state but lose individual values. Two blocks can therefore have identical retained discrete cosine transform (DCT) coefficients and different final errors. That endpoint is the latest observed error when the next nonoverlapping block is forecast. We test whether retaining it alongside the compressed summary and current base forecast improves correction.

We propose Endpoint-Preserving Online Correction (EPOC) for a fixed multi-horizon forecaster. EPOC retains low-order DCT coefficients of the preceding completed base-residual block and its uncompressed final value. For each channel, it supplies that endpoint as a shared feature to separate online ridge regressions for the retained correction components. Each regression also receives the corresponding coefficient of the current base forecast. The fitted components reconstruct a correction in the same fixed DCT subspace, and a blending weight derived from completed blocks combines it with the base forecast. Updates occur after the full target block has arrived; the retained auxiliary arrays include regression statistics, the residual summary, the fixed DCT basis, and blending state.

We evaluate three questions. First, how do correction accuracy and retained auxiliary arrays compare with other online methods? Second, what do the endpoint and the other input components contribute at a fixed output dimension? Third, how does the number of retained components affect the accuracy--storage choice? The main comparison applies the uncorrected base, five correction comparators, and EPOC to eight multivariate series, DLinear and PatchTST bases, three seeds, and two base-training variants, yielding 96 matched fixed-base conditions. Equal-size first, middle, and mean residual summaries isolate the position and type of retained scalar; input-removal and projected-endpoint controls examine the complete regression input. A separate experiment compares the correction with TEFL-style residual adapters on the same jointly trained bases.

Across the 96 fixed-base conditions, EPOC reduces mean squared error (MSE) by an average of 15.40\% and mean absolute error (MAE) by 9.35\% relative to the uncorrected base, while retaining a median of 6,352 B of auxiliary numerical arrays. The endpoint yields 1.65--2.20\% lower paired MSE than the three equal-size residual summaries. Full ELF achieves a larger mean MSE reduction from the uncorrected base, 19.29\%, with 474,048 median retained bytes. EPOC retains about one seventy-fifth of ELF's auxiliary state, with a 3.89-percentage-point smaller mean MSE reduction. On jointly trained bases, it lowers MSE by 16.69--20.15\% relative to the globally blended TEFL-style adapter applied to the same base.

The contributions are as follows:
\begin{enumerate}
\item We define a compact retained state that pairs truncated residual DCT coefficients with an uncompressed endpoint and uses current-forecast coefficients in channel- and component-wise online correction.
\item We evaluate the retained-information choice through equal-size residual-summary controls, a coefficient-reconstructed endpoint, and input ablations under matched base forecasts and correction dimensions.
\item We compare accuracy and retained arrays over 96 fixed-base conditions, jointly trained adapters, and component and update settings.
\end{enumerate}

Section~\ref{sec:related_work} reviews related correction and adaptation methods. Section~\ref{sec:method} defines the forecast--feedback order and retained state; Section~\ref{sec:experimental_setup} gives the evaluation protocol. Section~\ref{sec:results} presents the results, followed by discussion and conclusions in Sections~\ref{sec:discussion} and~\ref{sec:conclusion}.

\section{Related Work}
\label{sec:related_work}

\subsection{Residual feedback for multi-horizon forecasts}
Prediction errors from completed forecasts can inform the next correction. ResCAL~\cite{rescal2022} estimates future traffic-forecasting errors from previously observed residuals and graph signals. TEFL~\cite{tefl2026} addresses the timing of residual feedback in rolling multi-horizon forecasts: it selects the most recent fully observable residual vector, maps that vector through a low-rank additive adapter, and jointly trains the adapter and base forecaster after a base-model warmup. TEFL's complete residual input already includes its final value. We examine how to retain that value explicitly when the residual block is compressed.

EPOC addresses the retained representation of a completed block. It keeps a small set of fixed DCT coefficients together with the uncompressed final residual, shares that endpoint across component-wise regressions, and also uses coefficients of the current base forecast. The base forecaster remains fixed while the regression statistics update after each completed block. The joint-base comparison in Table~\ref{tab:learned_comparison} evaluates this correction against a TEFL-style residual adapter; the equal-size summary in Table~\ref{tab:endpoint_selection} and input controls in Fig.~\ref{fig:input_ablation} examine the retained endpoint and its combination with compressed features.

\subsection{Calibration of fixed forecasters}
Methods that retain a fixed base differ in the information supplied to their auxiliary predictors and in what they update. ELF~\cite{elf2025} learns an online forecast from Fourier features of observed history and combines it with the fixed model's forecast through an adaptive weight. TAFAS~\cite{tafas2025} places gated temporal calibration modules around the base and can adapt using partially observed targets. PETSA~\cite{petsa2025} uses low-rank input and output calibration modules with robust, spectral, and patch-structure loss terms. The $\delta$-Adapter~\cite{deltaadapter2026} attaches bounded input and output adjustments to a deployed forecaster, including an additive output correction.

COSA~\cite{cosa2026} applies a gated linear correction to the current forecast using statistics of recently observed targets as context. FAC~\cite{fac2026} calibrates the input and output through gated complex-affine transformations in the frequency domain, updating from matured targets. OMPB~\cite{ompb2026} trains a gated Bayesian residual head on a fixed forecast under an online objective that includes source-risk and source-to-target mismatch terms. These methods establish several routes to online correction: a separate forecast, input/output calibration, or an output-side adapter. Our correction uses the preceding forecast's residual coefficients and endpoint as explicit inputs, with the current forecast coefficients, and constrains its output to a fixed DCT subspace. Table~\ref{tab:prior_methods} compares the auxiliary information, feedback mechanism, and retained state of the methods evaluated in this study.

\subsection{Sequential adaptation and retained information}
Other online forecasters adapt the forecasting system or regulate when it should change. FSNet~\cite{fsnet2023} combines fast adaptation with memory for recurring patterns, and OneNet~\cite{onenet2023} updates and combines models of temporal and cross-variable dependencies. Reconditionor identifies context-driven shifts through residual--context dependence, while SOLID selects similar samples to fine-tune a prediction layer~\cite{solid2024}. DynaTTA~\cite{dynatta2025} adjusts adaptation using prediction-error and representation-drift signals; PADRE~\cite{padre2026} uses delayed residual trajectories and retrieved drift patterns to gate backbone updates. EPOC keeps the backbone fixed and uses completed residuals directly as features for an auxiliary correction.

The DCT~\cite{ahmed1974dct}, ridge regression~\cite{hoerl1970ridge}, and forgetting in sequential least squares~\cite{goel2020rls} provide established tools for compressing and estimating from a stream. The design choice examined here is which information from a completed residual block to keep alongside its compressed coefficients. The endpoint, residual coefficients, and current-forecast coefficients define the regression input; the fixed output subspace and sufficient statistics determine the retained state analyzed in Section~\ref{sec:method}.

\begin{table}[t]
\caption{Auxiliary inputs, correction mechanisms, and retained state in the evaluated comparison methods. Existing methods are ordered by publication year; EPOC appears last.}
\label{tab:prior_methods}
\centering
% Focused structural comparison; numerical transfers are specified in Section IV.
\begingroup\fontsize{9}{11}\selectfont
\setlength{\tabcolsep}{3pt}
\begin{tabular}{@{}>{\raggedright\arraybackslash}p{.20\linewidth}>{\raggedright\arraybackslash}p{.24\linewidth}>{\raggedright\arraybackslash}p{.28\linewidth}>{\raggedright\arraybackslash}p{.20\linewidth}@{}}
\toprule
Method & Auxiliary input & Correction and feedback & Retained state \\
\midrule
ELF (ICML 2025) & Fourier features of observed history & Online auxiliary forecast and adaptive mixing & Regression and mixing statistics \\[3pt]
TEFL (arXiv 2026) & Most recent completed residual block & Low-rank residual adapter; joint base training & Adapter weights and residual input \\[3pt]
$\delta$-Adapter (ICLR 2026) & Current input and forecast & Bounded input/output adapters; online updates & Adapter weights and optimizer state \\[3pt]
COSA (ICLR 2026) & Current forecast and recent target statistics & Gated linear output correction; online updates & Linear adapter, gate, context \\[3pt]
FAC (arXiv 2026) & Frequency representations of input and forecast & Gated frequency-domain calibration & Masks, biases, optimizer state \\[3pt]
OMPB (UAI 2026) & Base forecast & Gated Bayesian residual head; predict then update & Head, optimizer, source/target buffers \\[3pt]
EPOC & Past residual DCT coefficients and endpoint; current forecast coefficients & Fixed DCT output subspace; complete-block update & Ridge statistics, endpoint, mixing state \\
\bottomrule
\end{tabular}
\endgroup

\end{table}

\section{EPOC: Endpoint-Preserving Online Correction}
\label{sec:method}

\subsection{Forecasting and feedback order}
Let $t=1,2,\ldots$ index consecutive, nonoverlapping forecast blocks. At the origin of block $t$, a fixed base forecaster maps the preceding $L$ observations to $f_t\in\mathbb{R}^{H\times C}$, where $H$ is the forecast horizon and $C$ is the number of channels. The target $y_t\in\mathbb{R}^{H\times C}$ becomes available only after the block is complete. We define its base residual as
\begin{equation}
 e_t=y_t-f_t.
 \label{eq:base_residual}
\end{equation}
This residual is measured against the uncorrected base forecast, so its definition is independent of the correction issued for block $t$.

\begin{figure}[t]
\centering
\includegraphics[width=\textwidth]{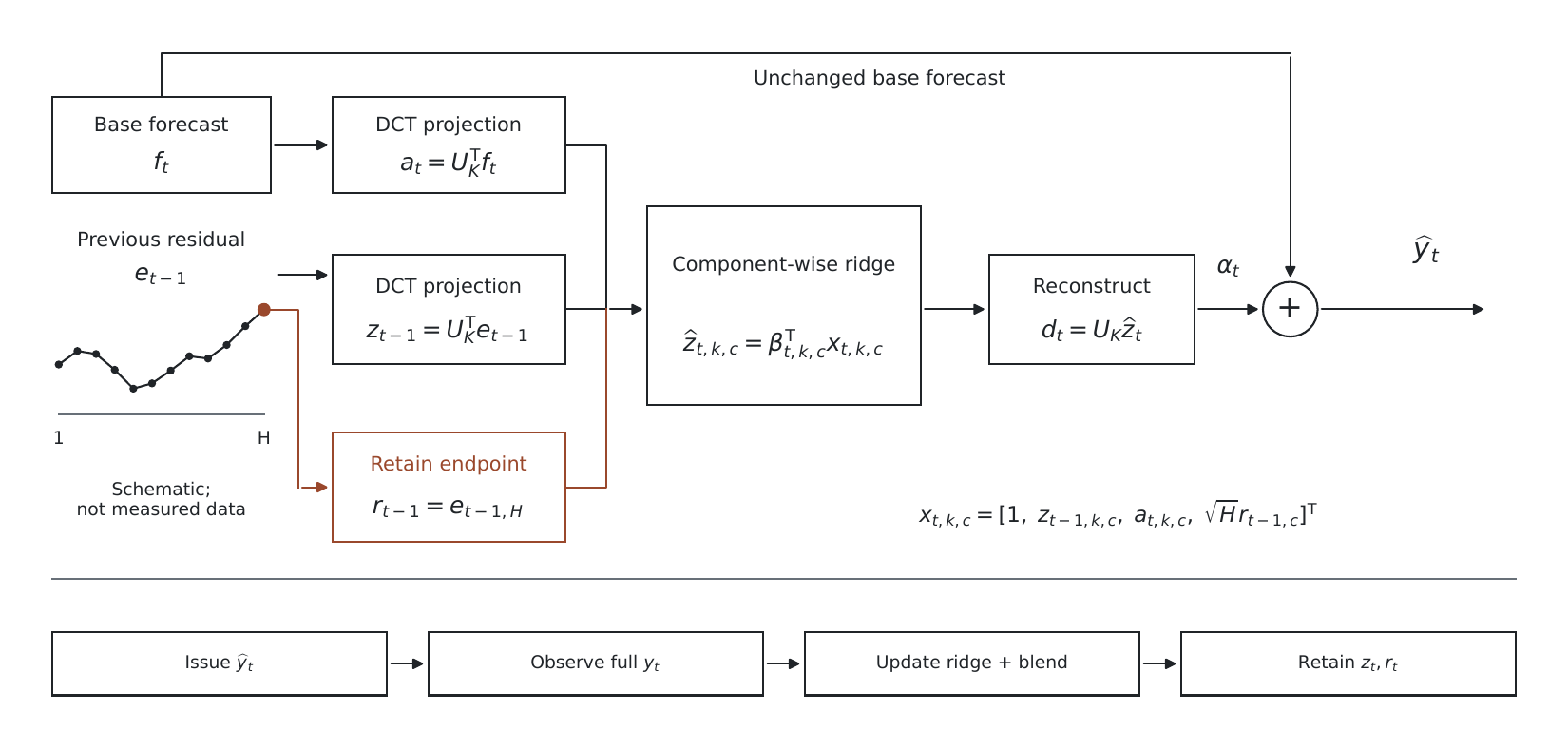}
\caption{Correction of block $t$ and the complete-block feedback order. The DCT coefficients and final value of $e_{t-1}$ are available when $f_t$ is issued. The residual trace is schematic. After $y_t$ is fully observed, its base residual updates the regression, the blending controller, and the summary used at the next origin.}
\label{fig:method}
\end{figure}

Forecast origins advance by $H$ samples. Consequently, $e_{t-1}$ is known when block $t$ is issued, while $e_t$ is not. The correction and its blending weight use only completed blocks. After all of $y_t$ has arrived, the method updates its regression and blending states and retains a summary of $e_t$ for block $t+1$. The experiments use $L=96$ and $H=24$. Figure~\ref{fig:method} shows the information flow and the order of issuance and feedback.

\subsection{Residual representation and online correction}
Let $U_K\in\mathbb{R}^{H\times K}$ contain the first $K$ columns of a fixed orthonormal DCT basis~\cite{ahmed1974dct}. The first column is constant. With horizon index $h=1,\ldots,H$ and component index $k=1,\ldots,K$,
\begin{equation}
 (U_K)_{h,k}=\begin{cases}
 H^{-1/2}, & k=1,\\
 \sqrt{2/H}\cos\!\left[\pi(h-\tfrac12)(k-1)/H\right], & k>1.
 \end{cases}
 \label{eq:dct_basis}
\end{equation}
Thus $U_K^{\mathsf T}U_K=I_K$. For channel $c$, we retain the coefficients of the preceding base residual and its observed final value, and transform the current base forecast:
\begin{equation}
 \begin{split}
 z_{t-1,c}&=U_K^{\mathsf T}e_{t-1,c},\qquad
 a_{t,c}=U_K^{\mathsf T}f_{t,c},\\
 r_{t-1,c}&=e_{t-1,H,c}.
 \end{split}
 \label{eq:retained_features}
\end{equation}
The reference configuration uses $K=4$. Before the first evaluated block, $z_{0,c}$ and $r_{0,c}$ are zero.

We estimate each correction component with a separate regression for each channel. Its four-dimensional input is
\begin{equation}
 x_{t,k,c}=\bigl[1,\ z_{t-1,k,c},\ a_{t,k,c},\
 \sqrt{H}\,r_{t-1,c}\bigr]^{\mathsf T}.
 \label{eq:regression_input}
\end{equation}
The same uncompressed endpoint $r_{t-1,c}$ is supplied to all $K$ regressions of channel $c$; it is stored once per channel. The factor $\sqrt H$ is fixed. If the residual is constant throughout a block, the scaled endpoint equals its constant DCT coefficient, so the two inputs have the same scale in that case.

The endpoint can distinguish residual blocks that have the same retained DCT coefficients. Let $\delta_H\in\mathbb{R}^H$ select the final sample and set $v=(I_H-U_KU_K^{\mathsf T})\delta_H$. When $K<H$, $U_K^{\mathsf T}v=0$, whereas $\delta_H^{\mathsf T}v=\lVert v\rVert_2^2>0$. Adding $v$ to a residual block therefore leaves its first $K$ coefficients unchanged but changes its endpoint. This establishes the information difference between the stored endpoint and the endpoint reconstructed from truncated coefficients; their forecasting effect is evaluated in Fig.~\ref{fig:input_ablation}.

For each $(k,c)$, let $\beta_{t,k,c}\in\mathbb{R}^4$ be the coefficients fitted before block $t$. We use exponentially weighted ridge regression~\cite{hoerl1970ridge,goel2020rls} on inputs recorded at their original forecast times:
\begin{equation}
 \begin{split}
 \beta_{t,k,c}=\mathop{\arg\min}_{b\in\mathbb{R}^4}\;&
 \lambda\lVert b\rVert_2^2\\
 &+\sum_{s=1}^{t-1}\rho^{t-1-s}
 \bigl(z_{s,k,c}-b^{\mathsf T}x_{s,k,c}\bigr)^2.
 \end{split}
 \label{eq:ridge_objective}
\end{equation}
The target $z_{s,k,c}$ is the coefficient of the base residual revealed after block $s$. The penalty includes the intercept. We set $\lambda=1$ and $\rho=2^{-1/128}$, giving a weight half-life of 128 completed blocks.

The objective can be updated without retaining previous blocks. With $(k,c)$ suppressed, initialize $G_1=\lambda I_4$ and $q_1=0$. After $y_t$ is observed, update
\begin{equation}
 \begin{split}
 G_{t+1}&=\rho G_t+x_t x_t^{\mathsf T}
 +(1-\rho)\lambda I_4,\\
 q_{t+1}&=\rho q_t+x_t z_t.
 \end{split}
 \label{eq:ridge_update}
\end{equation}
Solving $G_t\beta_t=q_t$ gives the minimizer of~\eqref{eq:ridge_objective}. The final term in the Gram update keeps the ridge penalty fixed while older observations lose weight. The implementation packs the symmetric statistics and solves the linear system directly.

The fitted coefficient, raw correction, and final issued forecast are
\begin{equation}
 \widehat z_{t,k,c}=\beta_{t,k,c}^{\mathsf T}x_{t,k,c},
 \qquad d_t=U_K\widehat z_t,
 \label{eq:raw_correction}
\end{equation}
\begin{equation}
 \widehat y_t=f_t+\alpha_t d_t.
 \label{eq:final_forecast}
\end{equation}
The correction for each channel lies in $\operatorname{span}(U_K)$. The separately retained endpoint changes the regression inputs, while $K$ fixes the correction output dimension.

The scalar $\alpha_t\in[0,1]$ is shared across all horizons and channels. For each completed block $s$, the controller records the squared magnitude of the raw correction and its agreement with the base residual:
\begin{equation}
 A_s=\frac{\lVert d_s\rVert_F^2}{HC},\qquad
 B_s=\frac{\langle d_s,e_s\rangle_F}{HC}.
 \label{eq:blend_statistics}
\end{equation}
Let $\mathcal W_t=\{\max(1,t-W),\ldots,t-1\}$ contain at most $W$ completed blocks. Minimizing the squared error of $f_s+\alpha d_s$ over these blocks for one bounded weight gives
\begin{equation}
 \alpha_t=\begin{cases}
 \operatorname{clip}_{[0,1]}\!\left(
 \dfrac{\sum_{s\in\mathcal W_t}B_s}
 {\sum_{s\in\mathcal W_t}A_s}\right),
 &\sum_{s\in\mathcal W_t}A_s>0,\\
 0,&\text{otherwise}.
 \end{cases}
 \label{eq:blend_weight}
\end{equation}
We use $W=32$. The controller uses the raw correction computed when block $s$ was issued; it does not recompute that correction with later regression coefficients. Algorithm~\ref{alg:online_correction} specifies the issue and update order.

\begin{algorithm}[H]
\caption{Online retained-state correction}
\label{alg:online_correction}
\renewcommand{\algorithmicrequire}{\textbf{Input:}}
\renewcommand{\algorithmicensure}{\textbf{Output:}}
\begin{algorithmic}[1]
\REQUIRE Base forecast $f_t$ at each origin; completed target $y_t$ after issuance; fixed $U_K$, $\lambda$, $\rho$, and $W$.
\ENSURE Corrected forecast $\widehat y_t$ issued before $y_t$ arrives.
\STATE Initialize $z_{0,c}=r_{0,c}=0$, $G_1=\lambda I_4$, $q_1=0$, and an empty controller.
\FOR{each block $t$}
\STATE Receive $f_t$; compute $a_t$, $x_{t,k,c}$, $\widehat z_t$, and $\alpha_t$ from prior state.
\STATE Issue $\widehat y_t=f_t+\alpha_t U_K\widehat z_t$; retain $d_t$ and $x_{t,k,c}$ until feedback.
\STATE After all of $y_t$ arrives, compute $e_t=y_t-f_t$ and $z_t=U_K^{\mathsf T}e_t$.
\STATE Update the controller with $(d_t,e_t)$ and each regression with $(x_{t,k,c},z_{t,k,c})$.
\STATE Retain $z_{t,c}$ and $r_{t,c}=e_{t,H,c}$ for block $t+1$.
\ENDFOR
\end{algorithmic}
\end{algorithm}

\subsection{Retained-state accounting}
We count numerical arrays kept by the correction method between forecast blocks. Its regression statistics contain, for each of the $KC$ regressions, six packed entries of the symmetric Gram matrix for three non-intercept features, three intercept--feature cross terms, and four target cross terms. The intercept weight sum is common to every regression and is stored once. These statistics therefore occupy $13KC+1$ floating-point values.

The other retained arrays contain $HK$ values for $U_K$, $KC$ previous residual coefficients, $C$ endpoints, and $2W+2$ values for the circular blending history and its running sums. The implementation also retains $K$ component indices. All counted values and indices use 64 bits, giving
\begin{equation}
 S(H,K,C,W)=8\bigl(HK+K+14KC+C+2W+3\bigr)
 \ \text{bytes}.
 \label{eq:state_bytes}
\end{equation}
At $H=24$, $K=4$, and $W=32$, this is $8(167+57C)$ bytes: 4,528 B for a seven-channel transformer-temperature series, 8,176 B for a 15-channel building-meter site, and 13,192 B for the 26-channel Appliances series. The median over the 96 fixed-base conditions is 6,352 B, as reported in Table~\ref{tab:main_results}. This is the size of persistent auxiliary numerical arrays. The base forecaster, common observation history, forecast and target buffers, temporary features and solver workspace, and language-runtime overhead are outside this count; it is not a measurement of peak random-access memory (RAM) or execution time.

For each block, projecting the base forecast and completed residual onto the $K$ DCT components and reconstructing the correction each cost $O(HKC)$ arithmetic operations. Updating the $KC$ four-feature ridge systems costs $O(KC\cdot 4^2)$, and solving them directly costs $O(KC\cdot 4^3)$. These counts describe the correction only; they exclude the base forecaster and are not measured latency or energy use.

\section{Experimental Setup}
\label{sec:experimental_setup}

\subsection{Data and common evaluation protocol}
Table~\ref{tab:datasets} lists eight multivariate series: ETTh1, ETTh2, ETTm1, and ETTm2 from the Electricity Transformer Temperature (ETT) data~\cite{ettdata}; Appliances Energy Prediction~\cite{appliancesdata}; and the Rat, Fox, and Panther sites from Building Data Genome 2 (BDG2)~\cite{bdg2data}. Each ETT series has seven channels. For Appliances, we omit the timestamp and the random variables \texttt{rv1} and \texttt{rv2}, leaving 26 numerical channels. Each BDG2 site contains 15 selected electricity meters. Forecasts and losses include every retained channel.

The BDG2 meter IDs and channel order are fixed in the released inputs. Meters were chosen by their missing-value fractions over the supplied series. Fox and Panther were forward filled and then backward filled from the pinned cleaned source. The archived Rat input differs from that procedure at 78 cells; explicit snapshot adjustments reproduce its evaluated values.

For every series, the lookback is $L=96$ samples, the horizon is $H=24$ samples, and consecutive forecast origins are 24 samples apart. These sample counts represent different durations: ETTh and BDG2 are hourly, ETTm is sampled every 15 minutes, and Appliances every 10 minutes. Let $T$ be the series length, $n=\lfloor T/2\rfloor$, and $a=\lfloor0.8n\rfloor$. The legacy base is fitted using $[0,a)$, with $[a,n)$ reserved for model and comparator selection. A refit base uses $[0,n)$ after its settings have been selected. Evaluation origins are $n,n+H,\ldots$; only complete target blocks within $[n,T)$ are scored. Per-channel means and standard deviations are computed on $[0,n)$; we subtract the mean and divide by the standard deviation plus $10^{-8}$. All reported errors use this normalized scale.

\begin{table}[!t]
\caption{Prepared evaluation series. Eval. is the zero-based index of the first evaluation target; Blocks counts complete 24-sample targets. All eight series enter the same fixed-base comparison.}
\label{tab:datasets}
\centering
% Dataset-specific quantities. Common L, H, and stride are defined in Section IV.
\begingroup\fontsize{9}{11}\selectfont
\setlength{\tabcolsep}{2.3pt}
\begin{tabular}{lrrrr}
\toprule
Series & Samples & $C$ & Eval. & Blocks \\
\midrule
ETTh1 & 17420 & 7 & 8710 & 362 \\
ETTh2 & 17420 & 7 & 8710 & 362 \\
ETTm1 & 69680 & 7 & 34840 & 1451 \\
ETTm2 & 69680 & 7 & 34840 & 1451 \\
Appliances & 19735 & 26 & 9867 & 411 \\
BDG2 (Rat) & 17544 & 15 & 8772 & 365 \\
BDG2 (Fox) & 17544 & 15 & 8772 & 365 \\
BDG2 (Panther) & 17544 & 15 & 8772 & 365 \\
\bottomrule
\end{tabular}
\endgroup

\end{table}

The fixed-base comparison uses DLinear~\cite{dlinear2022} and a compact, channel-independent PatchTST~\cite{patchtst2023}. DLinear decomposes the input with a replicate-padded moving average of width 25 and uses separate trend and remainder maps shared across channels. PatchTST uses length-16 patches at stride 8, a 64-dimensional embedding, four attention heads, two encoder layers, a feed-forward width of 128, zero dropout, a learned position embedding, and a linear forecast head. It also uses a learned affine input transformation per channel.

For each series, architecture, and seed in $\{0,1,2\}$, we train the legacy base with Adam at $10^{-3}$, batch size 32, and MSE loss on randomly sampled training windows. The update count is chosen by validation MSE from $\{200,500,1000,2000,4000,8000,20000\}$. The refit base is reinitialized with the corresponding seed and trained on $[0,n)$ for that selected count. Both models are frozen during evaluation. The eight series, two architectures, three seeds, and two training variants yield $8\times2\times3\times2=96$ conditions. Within each condition, all methods receive the same base forecasts, observed targets, and forecast origins. Each method issues a complete forecast before the target arrives and updates only after the full target block is observed. The first evaluation block is included in every reported loss.

\subsection{Comparators and learned comparison design}
Table~\ref{tab:prior_methods} distinguishes the auxiliary inputs and retained states of related methods. The main fixed-base comparison contains seven arms: the uncorrected base (Static), the input/output $\delta$-Adapter~\cite{deltaadapter2026}, COSA~\cite{cosa2026}, FAC~\cite{fac2026}, OMPB~\cite{ompb2026}, full ELF~\cite{elf2025}, and EPOC with $K=4$. All six correction arms use the same bounded, global, 32-completed-block blending rule of Section~\ref{sec:method}. Online states start empty at the evaluation boundary; weights trained before evaluation are retained. The base remains fixed, including when FAC calibrates its input.

The $\delta$-Adapter has bounded additive input and output multilayer perceptrons of width 512. The additive bound is 0.01 on ETT and 0.1 on the other series. For each fixed base, the adapter receives ten source-training epochs with Adam on at most 1024 sampled pre-validation windows. It takes one clipped Adam step when each evaluation target completes. Its online rate is selected from $\{10^{-6},10^{-4},10^{-3}\}$ using chronological online MSE on legacy validation origins starting at $a+96$. COSA uses the official linear adapter and adaptive rate function from commit \texttt{527c0feb}, with channel-specific maps and gates and ten entries of completed-block context. FAC is a reconstruction of the published input/output frequency-calibration equations using channel-specific complex masks and biases, inverse real fast Fourier transform (FFT), and a gated correction. COSA and FAC take three gradient steps per completed block. Their initial rates are selected on the same legacy validation origins from $\{10^{-4},10^{-3},5\times10^{-3},10^{-2}\}$. Each selection uses that series, architecture, and seed; the chosen rate is carried to the corresponding refit condition, and adapter state is reset before evaluation.

OMPB uses a gated per-channel Gaussian residual head with source-risk, divergence, and supervised-feedback terms. Its transferred settings are $K=5$, five optimizer steps per completed block, Adam rate 0.005, a 256-pair source buffer, and a 64-forecast target buffer. The head is pretrained for 20 epochs using at most 1024 source windows per fixed base; no validation search changes these settings. Full ELF uses cumulative Fourier-coordinate ridge regression with one newly completed training window per block, no pre-evaluation regression pairs, and ridge coefficient 20. Its input/output frequency counts are $(d,q)=(87,10)$. The input is centered by its channel mean before orthonormal Fourier transformation; the mean is restored after reconstruction. In unscaled sufficient statistics, the channel-$c$ ridge term is $20\sigma_c^2 I$. Before fitting, the auxiliary forecast repeats the most recent daily cycle. We retain independent real Fourier coordinates and packed symmetric statistics for storage accounting. ELF's original dynamic weighting is replaced by the shared global controller in the main comparison.

A separate experiment compares EPOC with the two-projection TEFL-style residual adapter~\cite{tefl2026}. It covers six series (the four ETT series, Appliances, and Rat), both architectures, and three seeds. The adapter has shared weights across channels, a ReLU hidden layer, no projection biases, and widths 2, 4, or 64; its second projection is initialized to zero. Base warmup lasts three chronological epochs under either MAE alone or MAE plus spectral flatness (SF) with unit weight. SF is the geometric-to-arithmetic mean ratio of full Fourier power, stabilized by $10^{-8}$, after consecutive residual blocks in a batch are concatenated in time and averaged over channels. The second stage uses 12 epochs of MAE optimization, either jointly for base and adapter or for the base alone. Both stages use AdamW at $10^{-3}$, weight decay 0.01, and batch size 32. Checkpoints are selected by validation MAE over epochs 0--12, with earlier epochs winning ties. Paired runs share the warmup state and shuffle seed.

At evaluation, a learned adapter receives the previous completed base residual and has fixed weights. We score its raw and globally blended forecasts. For each warmup and width, the direct EPOC-versus-adapter comparison applies both corrections to the \emph{same jointly trained base} with the same blending rule. A separately trained base-only run tests the benefit of the joint adapter procedure. We aggregate this group separately from the 96 fixed-base conditions.

The endpoint-location control replaces the most recent residual in EPOC's regression with the first residual, the twelfth residual, or the mean residual of the preceding block. Each scalar is multiplied by $\sqrt H$, enters all $K$ regressions of its channel, and has the same retained size as the endpoint. We run this control on all 96 fixed-base conditions and, separately, on the width-64 jointly trained bases under each warmup. For the fixed-base group, input ablations additionally remove the endpoint, the residual DCT coefficients, or the current forecast coefficients; compare endpoint-only regression and endpoint persistence; and supply an endpoint projected from the retained residual DCT coefficients. All fitted variants retain the same $K$-component output subspace. A 13-setting one-factor sweep changes $K$ to $1,2,3,6,8$, ridge strength to $0.1$ or $10$, forgetting half-life to $32,512$, or infinity completed blocks, and blending window to $8$ or $128$ blocks, alongside the reference setting. The input and setting studies each use all 96 fixed-base conditions.

\subsection{Metrics and aggregation}
For $N$ completed evaluation blocks, write $E_{t,h,c}=\widehat y_{t,h,c}-y_{t,h,c}$. We report normalized-scale errors
\begin{equation}
 \begin{aligned}
 \operatorname{MSE}&=\frac{1}{NHC}\sum_{t,h,c}E_{t,h,c}^{2},\\
 \operatorname{MAE}&=\frac{1}{NHC}\sum_{t,h,c}|E_{t,h,c}|.
 \end{aligned}
 \label{eq:eval_metrics}
\end{equation}
For metric $m\in\{\mathrm{MSE},\mathrm{MAE}\}$ and condition $i$, the reduction from Static is $100(1-m_{i,\mathrm{method}}/m_{i,\mathrm{Static}})$. A paired reduction from comparator $b$ to method $p$ instead uses $100(1-m_{i,p}/m_{i,b})$; positive values favor $p$. These are different denominators, and MSE and MAE are calculated separately.

Main overview values average the 96 condition-wise reductions and take the median of retained numerical-array bytes over those conditions. For Fig.~\ref{fig:accuracy_storage}, we first average the six condition-wise reductions for each series, backbone, and method, then average those series values within the ETT, BDG2, or Appliances group. Its retained-byte bar uses the common value for that method and channel group across series and backbones. In the jointly trained experiment, we first average each method's losses over three seeds within each of 12 series--architecture pairs, then compute the paired reduction and average those 12 percentages. Endpoint-location results keep the fixed-base and jointly trained groups separate. The order of aggregation matters: a mean of paired percentages generally differs from a percentage calculated from pooled losses.

\section{Results}
\label{sec:results}

\subsection{Accuracy and retained state}
Retained bytes count the correction parameters, optimizer tensors, regression statistics, residual summaries, method-specific buffers, and blending state kept between forecasts. The count excludes the frozen base, common observation history, forecast and target buffers, and temporary workspace.

Table~\ref{tab:main_results} compares seven methods on the 96 fixed-base conditions. EPOC reduces mean MSE by 15.40\% and mean MAE by 9.35\% relative to Static while retaining a median of 6,352 B. Its paired MSE reductions relative to the $\delta$-Adapter, COSA, FAC, and OMPB are 9.55\%, 4.18\%, 0.92\%, and 6.06\%, respectively. EPOC has lower MSE in 84, 77, 58, and 91 of the 96 paired conditions. All 96 EPOC conditions improve both MSE and MAE over Static. Full ELF has the highest mean reduction from Static, 19.29\% in MSE and 12.74\% in MAE, with 474,048 median retained bytes. EPOC therefore retains about one seventy-fifth of ELF's median auxiliary state, alongside a 3.89-percentage-point smaller mean MSE reduction from Static.

\begin{table}[!t]
\caption{Seven-method comparison on 96 matched conditions. The overview averages condition-wise reductions from Static; retained-byte multiples are relative to EPOC (6,352 B) and rounded to integers. The MSE and MAE panels average losses over three seeds and the legacy/refit variants within each series--backbone pair. The rightmost ratio divides EPOC's unrounded loss by the lowest unrounded loss in that row; 1 indicates the best result. Bold marks row-minimum losses and the corresponding ratio when EPOC is best. All corrections use global blending.}
\label{tab:main_results}
\centering
\begingroup\fontsize{9}{10.5}\selectfont
\setlength{\tabcolsep}{2.5pt}
\begin{tabular}{@{}lrrr@{}}
\toprule
\multicolumn{4}{@{}l}{\textit{96-condition overview}} \\
Method & MSE $\uparrow$ (\%) & MAE $\uparrow$ (\%) & Retained bytes $\downarrow$ \\
\midrule
Static & 0.00 & 0.00 & 0 ($\times0$) \\
$\delta$-Adapter & 6.70 & 5.12 & 22,552,608 ($\times3550$) \\
COSA & 11.73 & 7.56 & 111,756 ($\times18$) \\
FAC & 14.59 & 9.11 & 33,312 ($\times5$) \\
OMPB & 10.05 & 6.73 & 767,216 ($\times121$) \\
ELF & 19.29 & 12.74 & 474,048 ($\times75$) \\
EPOC & 15.40 & 9.35 & 6,352 ($\times1$) \\
\bottomrule
\end{tabular}
\par\vspace{5pt}
\begin{tabular}{@{}llrrrrrrrr@{}}
\toprule
\multicolumn{10}{@{}l}{\textit{MSE $\downarrow$}} \\
Dataset & Backbone & Static & $\delta$-Adapter & COSA & FAC & OMPB & ELF & EPOC & EPOC/best \\
\midrule
ETTh1 & DLinear & 0.3962 & 0.3939 & 0.3929 & 0.3840 & 0.3920 & 0.3673 & \textbf{0.3583} & \textbf{1.00000} \\
ETTh1 & PatchTST & 0.4378 & 0.4341 & 0.4143 & 0.3983 & 0.4018 & \textbf{0.3654} & 0.3779 & 1.03433 \\
ETTh2 & DLinear & 0.1453 & 0.1421 & 0.1397 & 0.1401 & 0.1414 & 0.1349 & \textbf{0.1341} & \textbf{1.00000} \\
ETTh2 & PatchTST & 0.1805 & 0.1752 & 0.1501 & 0.1407 & 0.1475 & \textbf{0.1382} & 0.1422 & 1.02896 \\
ETTm1 & DLinear & 0.2630 & 0.2570 & 0.2498 & \textbf{0.2366} & 0.2561 & 0.2390 & 0.2467 & 1.04298 \\
ETTm1 & PatchTST & 0.3167 & 0.3105 & 0.2653 & 0.2608 & 0.2793 & \textbf{0.2368} & 0.2718 & 1.14783 \\
ETTm2 & DLinear & 0.0998 & 0.0971 & 0.0960 & 0.0918 & 0.0965 & \textbf{0.0860} & 0.0895 & 1.04031 \\
ETTm2 & PatchTST & 0.1169 & 0.1120 & 0.0989 & 0.0943 & 0.0989 & \textbf{0.0862} & 0.0927 & 1.07540 \\
Appliances & DLinear & 0.2307 & 0.1994 & 0.2158 & 0.1858 & 0.2118 & 0.1932 & \textbf{0.1843} & \textbf{1.00000} \\
Appliances & PatchTST & 0.4393 & 0.3476 & 0.2018 & 0.1897 & 0.2059 & \textbf{0.1803} & 0.1827 & 1.01291 \\
BDG2 (Rat) & DLinear & 0.2193 & 0.2049 & 0.2011 & 0.1992 & 0.2062 & \textbf{0.1839} & 0.1929 & 1.04911 \\
BDG2 (Rat) & PatchTST & 0.2129 & 0.1941 & 0.1897 & 0.1857 & 0.1917 & \textbf{0.1737} & 0.1871 & 1.07681 \\
BDG2 (Fox) & DLinear & 0.2375 & 0.2024 & 0.1993 & 0.2065 & 0.2243 & \textbf{0.1761} & 0.2091 & 1.18712 \\
BDG2 (Fox) & PatchTST & 0.1817 & 0.1632 & 0.1623 & 0.1605 & 0.1689 & \textbf{0.1545} & 0.1714 & 1.10953 \\
BDG2 (Panther) & DLinear & 0.0779 & 0.0726 & 0.0715 & 0.0706 & 0.0757 & \textbf{0.0645} & 0.0666 & 1.03126 \\
BDG2 (Panther) & PatchTST & 0.0752 & 0.0689 & 0.0690 & 0.0670 & 0.0697 & \textbf{0.0642} & 0.0673 & 1.04854 \\
\bottomrule
\end{tabular}
\par\vspace{5pt}
\begin{tabular}{@{}llrrrrrrrr@{}}
\toprule
\multicolumn{10}{@{}l}{\textit{MAE $\downarrow$}} \\
Dataset & Backbone & Static & $\delta$-Adapter & COSA & FAC & OMPB & ELF & EPOC & EPOC/best \\
\midrule
ETTh1 & DLinear & 0.4040 & 0.4021 & 0.3989 & 0.3951 & 0.3984 & \textbf{0.3863} & 0.3863 & 1.00004 \\
ETTh1 & PatchTST & 0.4400 & 0.4364 & 0.4221 & 0.4122 & 0.4159 & \textbf{0.3922} & 0.4083 & 1.04097 \\
ETTh2 & DLinear & 0.2577 & 0.2538 & 0.2498 & 0.2513 & 0.2507 & 0.2443 & \textbf{0.2429} & \textbf{1.00000} \\
ETTh2 & PatchTST & 0.2974 & 0.2902 & 0.2633 & 0.2529 & 0.2596 & \textbf{0.2498} & 0.2555 & 1.02289 \\
ETTm1 & DLinear & 0.3389 & 0.3327 & 0.3263 & \textbf{0.3177} & 0.3313 & 0.3191 & 0.3266 & 1.02787 \\
ETTm1 & PatchTST & 0.3615 & 0.3555 & 0.3363 & 0.3339 & 0.3435 & \textbf{0.3182} & 0.3433 & 1.07887 \\
ETTm2 & DLinear & 0.2115 & 0.2072 & 0.2051 & 0.2009 & 0.2056 & \textbf{0.1915} & 0.1982 & 1.03468 \\
ETTm2 & PatchTST & 0.2329 & 0.2247 & 0.2097 & 0.2030 & 0.2088 & \textbf{0.1923} & 0.2030 & 1.05589 \\
Appliances & DLinear & 0.2870 & 0.2565 & 0.2788 & 0.2560 & 0.2752 & 0.2562 & \textbf{0.2491} & \textbf{1.00000} \\
Appliances & PatchTST & 0.3857 & 0.3346 & 0.2667 & 0.2604 & 0.2689 & \textbf{0.2493} & 0.2503 & 1.00408 \\
BDG2 (Rat) & DLinear & 0.3095 & 0.2869 & 0.2856 & 0.2854 & 0.2900 & \textbf{0.2659} & 0.2814 & 1.05813 \\
BDG2 (Rat) & PatchTST & 0.3140 & 0.2906 & 0.2850 & 0.2824 & 0.2876 & \textbf{0.2648} & 0.2860 & 1.08016 \\
BDG2 (Fox) & DLinear & 0.3171 & 0.2840 & 0.2888 & 0.2954 & 0.3042 & \textbf{0.2671} & 0.2993 & 1.12034 \\
BDG2 (Fox) & PatchTST & 0.2912 & 0.2698 & 0.2679 & 0.2679 & 0.2747 & \textbf{0.2568} & 0.2788 & 1.08569 \\
BDG2 (Panther) & DLinear & 0.1786 & 0.1704 & 0.1702 & 0.1692 & 0.1759 & \textbf{0.1598} & 0.1632 & 1.02153 \\
BDG2 (Panther) & PatchTST & 0.1770 & 0.1644 & 0.1655 & 0.1627 & 0.1661 & \textbf{0.1590} & 0.1647 & 1.03599 \\
\bottomrule
\end{tabular}
\endgroup

\end{table}

\begin{figure}[!t]
\centering
\includegraphics[width=.85\textwidth]{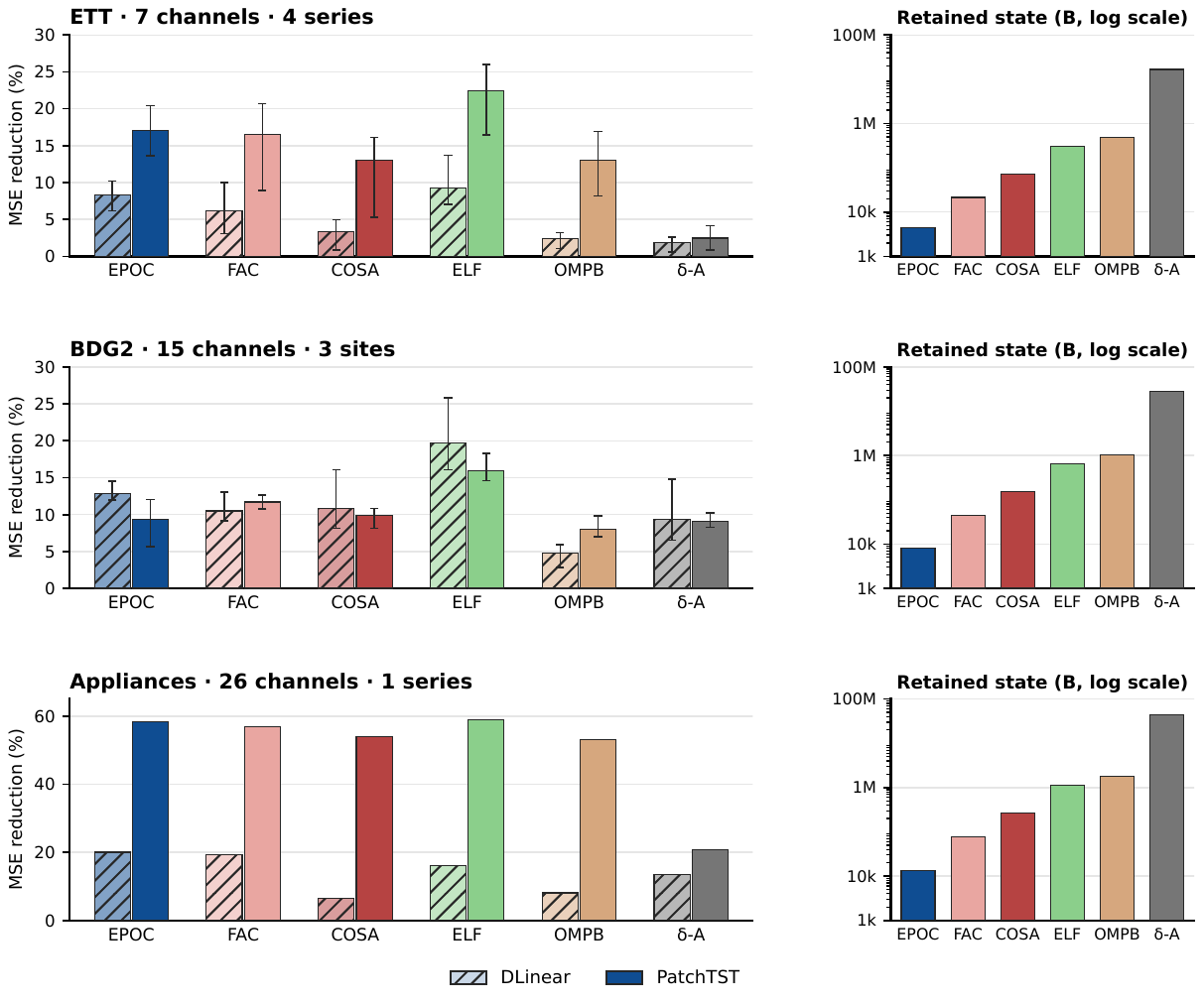}
\caption{Accuracy and retained correction state by channel group for six methods on 96 fixed-base conditions. Left bars show mean Static-relative MSE reduction: each series--backbone value averages three seeds and two training variants, and bars then average four ETT series, three BDG2 sites, or one Appliances series. Hatched bars denote DLinear and solid bars PatchTST. Whiskers show the range across series, not statistical uncertainty; Appliances has one series and no whiskers. Right bars show retained numerical-array bytes for each method and group on a logarithmic scale; this value is common to both backbones and all series in the group.}
\label{fig:accuracy_storage}
\end{figure}

Figure~\ref{fig:accuracy_storage} groups the eight series by channel count and places each method's mean Static-relative MSE reduction beside its retained-byte count. Separate DLinear and PatchTST accuracy bars show the backbone difference; the storage count is the same for both backbones within a group. EPOC has the smallest retained state among the six plotted corrections in all three groups. The group means can hide series-specific rankings: Table~\ref{tab:main_results} shows that COSA, FAC, and the $\delta$-Adapter each have lower MSE than EPOC on Fox with both backbones. ELF has the lowest MSE on both Fox and Panther backbones, with substantially more retained state than EPOC.

Table~\ref{tab:learned_comparison} evaluates EPOC against the TEFL-style residual adapter on jointly trained bases. Across the six warmup--width settings, EPOC lowers MSE by 16.69--20.15\% and MAE by 8.82--11.29\% relative to the globally blended adapter on the \emph{same} joint base. In a separate training comparison, the raw joint adapter lowers MSE by 2.22--6.44\% relative to its paired base-only run. Thus, the direct correction comparison is made against adapters that also improve on base-only training. EPOC retains 4,528 B in each joint-base comparison. The globally blended adapters retain median sizes of 1,584 B, 1,968 B, and 13,488 B at widths 2, 4, and 64: the small adapters use less state, while the width-64 adapter uses more.

\begin{table}[!t]
\caption{Joint-training comparison over six series, two backbones, and three seeds. The same-base columns give the paired reduction of EPOC relative to the globally blended adapter on the same jointly trained base. The separate-training columns give the reduction of the raw joint adapter relative to a separately trained base-only run. For each metric, seed losses are averaged within each of 12 series--backbone pairs before calculating reductions, which are then averaged across pairs. Retained bytes are medians for the globally blended corrections. SF denotes spectral flatness in base warmup; positive reductions favor the first-named method.}
\label{tab:learned_comparison}
\centering
\begingroup\fontsize{9}{11}\selectfont
\setlength{\tabcolsep}{4pt}
\renewcommand{\arraystretch}{1.15}
\begin{tabular}{@{}llrrrrrr@{}}
\toprule
& & \multicolumn{2}{c}{Same-base: EPOC vs. adapter} & \multicolumn{2}{c}{Separate training: adapter vs. base-only} & \multicolumn{2}{c}{Retained state (B)} \\
\cmidrule(lr){3-4}\cmidrule(lr){5-6}\cmidrule(l){7-8}
Warmup & Width & MSE (\%) & MAE (\%) & MSE (\%) & MAE (\%) & EPOC & Adapter \\
\midrule
MAE & 2 & 19.27 & 10.78 & 2.22 & 1.29 & 4,528 & 1,584 \\
& 4 & 18.98 & 10.61 & 3.13 & 1.77 & 4,528 & 1,968 \\
& 64 & 16.69 & 8.82 & 6.09 & 3.68 & 4,528 & 13,488 \\
\midrule
MAE+SF & 2 & 20.15 & 11.29 & 2.49 & 1.27 & 4,528 & 1,584 \\
& 4 & 19.54 & 10.86 & 3.68 & 1.98 & 4,528 & 1,968 \\
& 64 & 17.21 & 9.06 & 6.44 & 3.74 & 4,528 & 13,488 \\
\bottomrule
\end{tabular}
\endgroup

\end{table}

\subsection{Endpoint selection and input construction}
Table~\ref{tab:endpoint_selection} compares equal-size residual summaries in EPOC's regression. Over the 96 fixed-base conditions, the retained endpoint lowers mean paired MSE by 2.20\%, 1.65\%, and 1.96\% relative to the first residual, middle residual, and block mean, respectively. The corresponding MAE reductions are 1.46\%, 1.11\%, and 1.39\%. The joint-base groups show the same ordering of positive endpoint gains: their MSE reductions range from 4.70\% to 6.00\% with MAE warmup and from 5.43\% to 6.94\% with MAE+SF warmup. Within each condition, the endpoint and its three controls retain the same number of bytes, so these comparisons evaluate the chosen scalar summary at fixed regression and storage size.

\begin{table}[!t]
\centering
\caption{Endpoint gain over equal-size residual summaries. Each entry gives the mean paired reduction when the preceding block's endpoint replaces the listed first residual, middle residual, or block mean. The frozen-base row averages 96 condition-wise reductions; each joint row averages reductions across 12 series--backbone pairs after seed-mean losses are formed. All variants within a comparison use the same base, regression settings, blending rule, and retained array size.}
\label{tab:endpoint_selection}
\begingroup\fontsize{9}{11}\selectfont
\setlength{\tabcolsep}{5pt}
\renewcommand{\arraystretch}{1.18}
\begin{tabular}{@{}lrrrrrrr@{}}
\toprule
& \multicolumn{3}{c}{MSE reduction (\%)} & \multicolumn{3}{c}{MAE reduction (\%)} & \\
\cmidrule(lr){2-4}\cmidrule(lr){5-7}
Base / training & First & Middle & Mean & First & Middle & Mean & Retained (B) \\
\midrule
Frozen base (96 conditions) & 2.20 & 1.65 & 1.96 & 1.46 & 1.11 & 1.39 & 6,352 \\
Joint MAE (12 pairs) & 4.70 & 4.83 & 6.00 & 2.93 & 2.95 & 3.61 & 4,528 \\
Joint MAE+SF (12 pairs) & 5.43 & 5.49 & 6.94 & 3.38 & 3.34 & 4.14 & 4,528 \\
\bottomrule
\end{tabular}
\endgroup

\end{table}

\begin{figure}[!t]
\centering
\includegraphics[width=.75\textwidth]{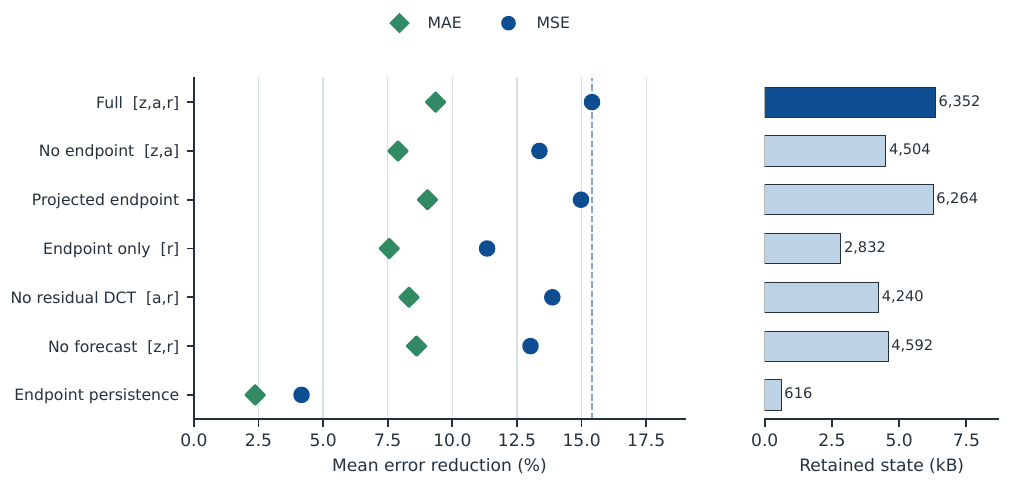}
\caption{Input construction on 96 matched fixed-base conditions. Markers show mean condition-wise MSE and MAE reductions from Static; bars show median retained bytes with global blending. The projected endpoint $\tilde r$ is reconstructed from retained residual DCT coefficients and supplied to every component regression. Endpoint persistence repeats the preceding endpoint without a fitted regression.}
\label{fig:input_ablation}
\end{figure}

Figure~\ref{fig:input_ablation} tests the complete regression input on the same 96 fixed-base conditions. The complete input yields a 15.40\% mean MSE reduction from Static, compared with 13.37\% without the endpoint, 13.87\% without residual DCT coefficients, 13.02\% without current forecast coefficients, and 11.34\% with the endpoint alone. In paired losses, the complete input reduces MSE by 4.55\% relative to endpoint-only regression and wins in all 96 conditions. An endpoint projected from the retained DCT coefficients yields a 14.98\% mean MSE reduction from Static, close to the 15.40\% of the observed-endpoint variant. Supplying that projected value to all component regressions therefore recovers much of the gain obtained by explicitly sharing an endpoint input. The complete method retains 6,352 median bytes, versus 6,264 B with the projected endpoint and 2,832 B with endpoint-only regression.

\subsection{Accuracy--storage and setting sensitivity}
\begin{figure}[H]
\centering
\includegraphics[width=.80\textwidth]{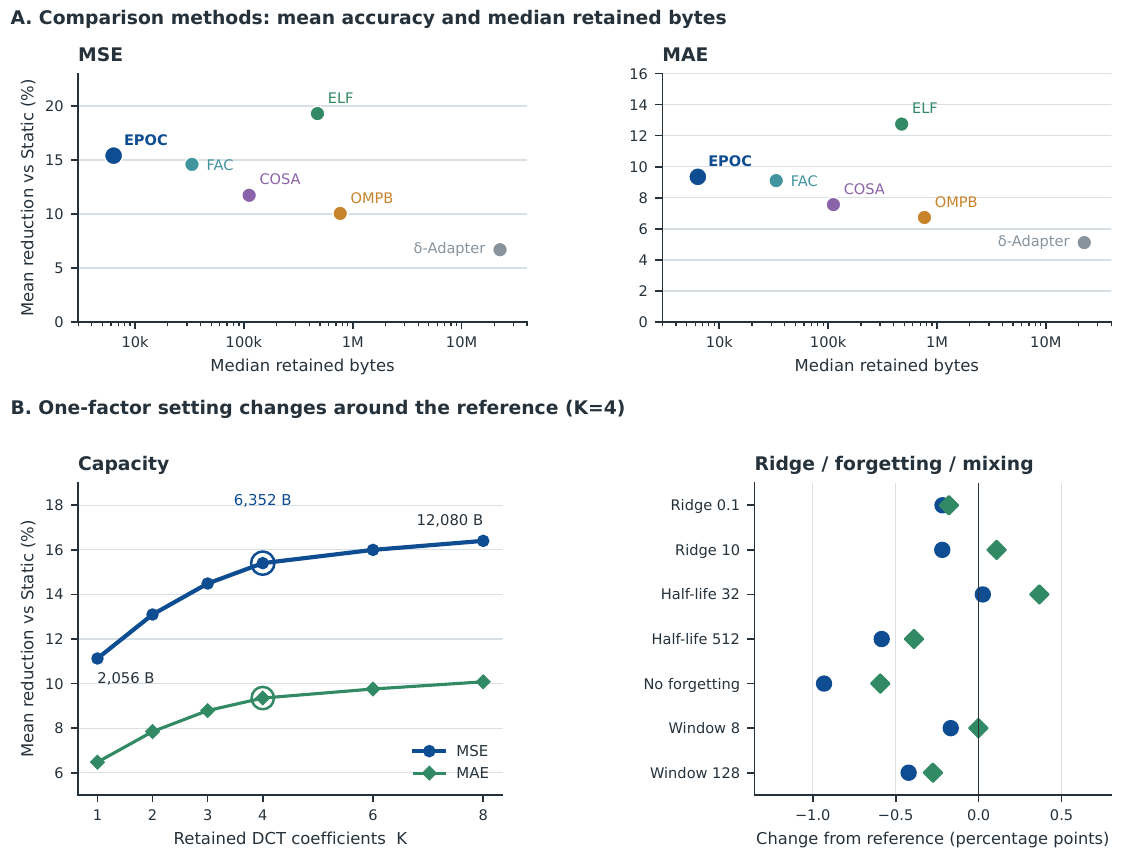}
\caption{Main comparison and one-factor sensitivity on 96 fixed-base conditions. Upper panels plot mean condition-wise reductions from Static against median retained bytes for the six globally blended correction methods. Lower left varies the retained DCT component count $K$; lower right changes ridge strength, forgetting half-life, or blending window relative to the $K=4$ reference. Unchanged settings follow Section~\ref{sec:method}.}
\label{fig:overview_sensitivity}
\end{figure}

The upper panels of Fig.~\ref{fig:overview_sensitivity} put all six corrections from Table~\ref{tab:main_results} on common MSE, MAE, and retained-state axes. EPOC gives a 15.40\% mean MSE reduction at 6,352 median bytes; full ELF gives 19.29\% at 474,048 B. FAC is the closest of the other main comparators in mean MSE reduction, at 14.59\% and 33,312 B. These points show the accuracy and storage choices behind the paired comparisons.

The lower panels vary one EPOC setting at a time over the same 96 conditions. Increasing $K$ from 1 to 4 raises the mean MSE reduction from 11.13\% to 15.40\% and median retained bytes from 2,056 to 6,352. At $K=8$, MSE reduction reaches 16.40\% with 12,080 B. The additional 1.00 percentage point from $K=4$ to $K=8$ requires 5,728 more retained bytes. With $K=4$ fixed, changing ridge strength, forgetting half-life, or blending window produces mean MSE reductions between 14.47\% and 15.43\%. Component count is the largest accuracy--storage choice among these one-factor changes. Table~\ref{tab:s_sensitivity} gives the full 13-setting summary, with condition-level values in the accompanying CSV.

\section{Discussion}
\label{sec:discussion}

The main comparison identifies an accuracy--storage operating point (Table~\ref{tab:main_results} and Fig.~\ref{fig:accuracy_storage}). On the 96 fixed-base conditions, EPOC reduces mean MSE by 15.40\% relative to Static with 6,352 median retained bytes. Full ELF increases that reduction to 19.29\%, but retains about 75 times as many bytes. EPOC also uses less state than the $\delta$-Adapter, COSA, FAC, and OMPB, and has lower paired MSE than each in a majority of conditions. The choice depends on the series and the available state budget: on BDG2 (Fox), the $\delta$-Adapter, COSA, and FAC have lower MSE with both backbones, while the width-2 and width-4 TEFL-style adapters in the separate joint-base experiment (Table~\ref{tab:learned_comparison}) retain fewer bytes than EPOC.

The endpoint controls clarify what the retained scalar contributes (Table~\ref{tab:endpoint_selection} and Fig.~\ref{fig:input_ablation}). At the same regression dimension and retained size, the preceding block's endpoint gives 1.65--2.20\% lower paired MSE than its first value, middle value, or mean. The low-order DCT coefficients summarize the completed block, whereas the observed endpoint explicitly records its latest residual; blocks with identical retained coefficients can have different endpoints. Without an endpoint input, the mean MSE reduction from Static falls to 13.37\%. Yet the endpoint projected from the retained coefficients achieves 14.98\%, close to the 15.40\% obtained with the observed endpoint. Much of the measured benefit therefore comes from supplying an endpoint-like feature to every component regression, with a smaller additional gain from retaining its uncompressed value in these experiments. The complete input also lowers paired MSE by 4.55\% relative to endpoint-only regression across the 96 conditions, supporting the combined use of residual coefficients, the shared endpoint feature, and current-forecast coefficients.

Among the tested settings, the number of retained components sets the principal accuracy--storage choice within EPOC (Fig.~\ref{fig:overview_sensitivity}). Increasing $K$ from 1 to 4 raises the mean MSE reduction from 11.13\% to 15.40\%, while increasing median retained state from 2,056 to 6,352 B. Increasing $K$ further to 8 adds 1.00 percentage point of MSE reduction and 5,728 B of state. Thus, $K=4$ is a compact reference setting rather than the most accurate tested setting; the preferred $K$ depends on the storage available for each channel.

The evidence covers eight series, two base forecasters, an input length of 96, a forecast horizon of 24, and updates after complete target blocks. Retained-byte counts describe persistent auxiliary numerical arrays between forecasts. Evaluating other horizons and series, and measuring peak memory, latency, and energy on target devices, would extend the accuracy--storage comparison to deployment costs.

\section{Conclusion}
\label{sec:conclusion}

This paper introduced EPOC, an endpoint-preserving online correction with compressed residual state for fixed multi-horizon forecasters. After each target block is complete, the method retains low-order DCT coefficients and the final value of the base residual, and shares that endpoint across component-wise online ridge regressions that also receive current-forecast coefficients. On 96 matched fixed-base conditions across eight series and two backbones, the complete correction yielded mean condition-wise MSE and MAE reductions from Static of 15.40\% and 9.35\%, respectively, while retaining a median of 6,352 B. It used less auxiliary state than the $\delta$-Adapter, COSA, FAC, and OMPB and yielded lower paired MSE than each in most conditions. Full ELF achieved a larger 19.29\% mean MSE reduction from Static with about 75 times the state. In the joint-base experiment, EPOC lowered MSE by 16.69--20.15\% relative to globally blended TEFL-style adapters applied to the same base. Equal-size residual-summary controls favored the endpoint by 1.65--2.20\% in paired MSE. The projected-endpoint and input-ablation controls indicate that sharing an endpoint-like feature across the component regressions, alongside residual and current-forecast coefficients, accounts for much of the measured gain. Increasing $K$ from 4 to 8 added 1.00 percentage point of mean MSE reduction at 5,728 B of additional retained state. These results identify a compact accuracy--storage operating point for online correction under complete-block feedback.

\bibliographystyle{unsrt}
\bibliography{refs}

@misc{tefl2026,
 author = {Huang, Xiannan and Fang, Shen and Qiu, Shuhan and Yu, Chengcheng and Du, Jiayuan and Yang, Chao},
 title = {{TEFL}: Prediction-Residual-Guided Rolling Forecasting for Multi-Horizon Time Series},
 year = {2026},
 howpublished = {arXiv:2602.22520, version 1},
 url = {https://arxiv.org/abs/2602.22520v1}
}

@misc{rescal2022,
 author = {Kim, Daejin and Cho, Youngin and Kim, Dongmin and Park, Cheonbok and Choo, Jaegul},
 title = {Residual Correction in Real-Time Traffic Forecasting},
 year = {2022},
 howpublished = {arXiv:2209.05406},
 url = {https://arxiv.org/abs/2209.05406v1}
}

@inproceedings{deltaadapter2026,
 author = {Liang, Daojun and Li, Qi and Wang, Yinglong and Chen, Jing and Zhang, Hu and Cui, Xiaoxiao and Wang, Qizheng and Li, Shuo},
 title = {The Forecast After the Forecast: A Post-Processing Shift in Time Series},
 booktitle = {International Conference on Learning Representations},
 year = {2026},
 url = {https://proceedings.iclr.cc/paper_files/paper/2026/hash/6f7d90b1198fec96defd80b5ebd5bc81-Abstract-Conference.html}
}

@inproceedings{elf2025,
 author = {Lee, Thomas L. and Toner, William and Singh, Rajkarn and Joosen, Artjom and Asenov, Martin},
 title = {Lightweight Online Adaption for Time Series Foundation Model Forecasts},
 booktitle = {Proceedings of the 42nd International Conference on Machine Learning},
 series = {Proceedings of Machine Learning Research},
 volume = {267},
 pages = {33736--33764},
 year = {2025},
 url = {https://proceedings.mlr.press/v267/lee25ag.html}
}

@article{ahmed1974dct,
 author = {Ahmed, N. and Natarajan, T. and Rao, K. R.},
 title = {Discrete Cosine Transform},
 journal = {IEEE Transactions on Computers},
 volume = {C-23},
 number = {1},
 pages = {90--93},
 year = {1974},
 doi = {10.1109/T-C.1974.223784}
}

@article{hoerl1970ridge,
 author = {Hoerl, Arthur E. and Kennard, Robert W.},
 title = {Ridge Regression: Biased Estimation for Nonorthogonal Problems},
 journal = {Technometrics},
 volume = {12},
 number = {1},
 pages = {55--67},
 year = {1970},
 doi = {10.2307/1267351}
}

@misc{ettdata,
 author = {Zhou, Haoyi},
 title = {{ETDataset}: Electricity Transformer Temperature Dataset},
 howpublished = {Dataset repository},
 url = {https://github.com/zhouhaoyi/ETDataset},
 note = {Accessed: Sep. 13, 2026}
}

@misc{appliancesdata,
 author = {Candanedo, Luis},
 title = {Appliances Energy Prediction},
 year = {2017},
 howpublished = {UCI Machine Learning Repository},
 doi = {10.24432/C5VC8G},
 url = {https://archive.ics.uci.edu/dataset/374/appliances+energy+prediction}
}

@article{bdg2data,
 author = {Miller, Clayton and Kathirgamanathan, Anjukan and Picchetti, Bianca and Arjunan, Pandarasamy and Park, June Young and Nagy, Zoltan and Raftery, Paul and Hobson, Brodie W. and Shi, Zixiao and Meggers, Forrest},
 title = {The Building Data Genome Project 2, energy meter data from the {ASHRAE} Great Energy Predictor {III} competition},
 journal = {Scientific Data},
 volume = {7},
 pages = {368},
 year = {2020},
 doi = {10.1038/s41597-020-00712-x}
}

@misc{dlinear2022,
 author = {Zeng, Ailing and Chen, Muxi and Zhang, Lei and Xu, Qiang},
 title = {Are Transformers Effective for Time Series Forecasting?},
 year = {2022},
 howpublished = {arXiv:2205.13504},
 url = {https://arxiv.org/abs/2205.13504}
}

@misc{patchtst2023,
 author = {Nie, Yuqi and Nguyen, Nam H. and Sinthong, Phanwadee and Kalagnanam, Jayant},
 title = {A Time Series is Worth 64 Words: Long-term Forecasting with Transformers},
 year = {2023},
 howpublished = {arXiv:2211.14730, version 2},
 url = {https://arxiv.org/abs/2211.14730v2}
}

@article{tafas2025,
 author = {Kim, HyunGi and Kim, Siwon and Mok, Jisoo and Yoon, Sungroh},
 title = {Battling the Non-stationarity in Time Series Forecasting via Test-time Adaptation},
 journal = {Proceedings of the AAAI Conference on Artificial Intelligence},
 volume = {39},
 number = {17},
 pages = {17868--17876},
 year = {2025},
 doi = {10.1609/aaai.v39i17.33965}
}

@misc{petsa2025,
 author = {Medeiros, Heitor R. and Sharifi-Noghabi, Hossein and Oliveira, Gabriel L. and Irandoust, Saghar},
 title = {Accurate Parameter-Efficient Test-Time Adaptation for Time Series Forecasting},
 year = {2025},
 howpublished = {arXiv:2506.23424, version 1},
 note = {Second Workshop on Test-Time Adaptation, ICML 2025},
 url = {https://arxiv.org/abs/2506.23424v1}
}

@misc{fac2026,
 author = {Wang, Haochun and Xu, Ruichen and Kementzidis, Georgios and Cho, Karen and Ramirez Villarreal, Sebastian and Deng, Yuefan},
 title = {Towards Principled Test-Time Adaptation for Time Series Forecasting},
 year = {2026},
 howpublished = {arXiv:2605.17250, version 1},
 url = {https://arxiv.org/abs/2605.17250v1}
}

@inproceedings{cosa2026,
 author = {Im, Jeonghwan and Kwon, Hyuk-Yoon},
 title = {{COSA}: Context-aware Output-Space Adapter for Test-Time Adaptation in Time Series Forecasting},
 booktitle = {International Conference on Learning Representations},
 year = {2026},
 url = {https://openreview.net/forum?id=L7Z5wBMPrW}
}

@inproceedings{solid2024,
 author = {Chen, Mouxiang and Shen, Lefei and Fu, Han and Li, Zhuo and Sun, Jianling and Liu, Chenghao},
 title = {Calibration of Time-Series Forecasting: Detecting and Adapting Context-Driven Distribution Shift},
 booktitle = {Proceedings of the 30th ACM SIGKDD Conference on Knowledge Discovery and Data Mining},
 pages = {341--352},
 year = {2024},
 doi = {10.1145/3637528.3671926},
 url = {https://arxiv.org/abs/2310.14838v2}
}

@inproceedings{padre2026,
 author = {Li, Xingwang and Teng, Fei and Zhou, Cong and Duan, Qiang},
 title = {Beyond Uniform Updates: Drift Pattern Aware Online Time Series Forecasting Under Delayed Feedback},
 booktitle = {Proceedings of the Thirty-Fifth International Joint Conference on Artificial Intelligence},
 pages = {4518--4526},
 year = {2026},
 doi = {10.24963/ijcai.2026/503},
 url = {https://www.ijcai.org/proceedings/2026/503}
}

@misc{dynatta2025,
 author = {Grover, Shivam and Etemad, Ali},
 title = {Shift-Aware Test-Time Adaptation and Benchmarking for Time-Series Forecasting},
 year = {2025},
 howpublished = {Second Workshop on Test-Time Adaptation: Putting Updates to the Test, ICML 2025},
 url = {https://openreview.net/forum?id=a399SmgWGl}
}

@inproceedings{ompb2026,
 author = {Huang, Chenfeng and Ma, Zixuan and Michailidis, George},
 title = {Model-Agnostic Online Certificate-Driven Calibration for Time Series Forecasting Under Distribution Shift},
 booktitle = {Proceedings of the 42nd Conference on Uncertainty in Artificial Intelligence},
 series = {Proceedings of Machine Learning Research},
 volume = {337},
 pages = {2244--2273},
 year = {2026},
 url = {https://proceedings.mlr.press/v337/huang26b.html}
}

@inproceedings{fsnet2023,
 author = {Pham, Quang and Liu, Chenghao and Sahoo, Doyen and Hoi, Steven C. H.},
 title = {Learning Fast and Slow for Online Time Series Forecasting},
 booktitle = {International Conference on Learning Representations},
 year = {2023},
 url = {https://openreview.net/forum?id=q-PbpHD3EOk}
}

@inproceedings{onenet2023,
 author = {Zhang, Yi-Fan and Wen, Qingsong and Wang, Xue and Chen, Weiqi and Sun, Liang and Zhang, Zhang and Wang, Liang and Jin, Rong and Tan, Tieniu},
 title = {{OneNet}: Enhancing Time Series Forecasting Models under Concept Drift by Online Ensembling},
 booktitle = {Advances in Neural Information Processing Systems},
 volume = {36},
 year = {2023},
 url = {https://proceedings.neurips.cc/paper_files/paper/2023/hash/dd6a47bc0aad6f34aa5e77706d90cdc4-Abstract.html}
}

@article{goel2020rls,
 author = {Goel, Ankit and Bruce, Adam L. and Bernstein, Dennis S.},
 title = {Recursive Least Squares with Variable-Direction Forgetting: Compensating for the Loss of Persistency},
 journal = {IEEE Control Systems Magazine},
 volume = {40},
 number = {4},
 pages = {80--102},
 year = {2020},
 doi = {10.1109/MCS.2020.2990516}
}

\appendix
% Number the supplementary tables from S1.
\setcounter{table}{0}
\renewcommand{\thetable}{S\arabic{table}}
\renewcommand{\theHtable}{supp.\arabic{table}}

\section{Supplementary Numerical Records}
\label{app:records}

\begin{table}[H]
\caption{Common evaluation settings and learned-adapter training configuration for the supplementary records. The fixed-base training and comparator protocols are given in Section~\ref{sec:experimental_setup}.}
\label{tab:s_settings}
\centering
{\fontsize{9}{11}\selectfont
\begin{tabular}{ll}
\toprule
Item & Setting \\
\midrule
Lookback / horizon / stride & 96 / 24 / 24 \\
EPOC residual basis & DCT, $K=4$ \\
Ridge / forgetting & $\lambda=1$, $\rho=2^{-1/128}$ \\
Blending window & 32 completed blocks \\
Learned widths / seeds & 2, 4, 64 / 0, 1, 2 \\
Warmup / joint training & 3 / 12 epochs \\
Joint loss / optimizer & MAE / AdamW \\
Learning rate / weight decay / batch & $10^{-3}$ / 0.01 / 32 \\
SF weight / power stabilization & 1 / $10^{-8}$ \\
ELF transfer ridge & $20\sigma_c^2 I$ in unscaled statistics; one window per block \\
Checkpoint selection & Validation MAE, epochs 0--12 \\
Storage excludes & Base model, observation history, forecast/target buffers,\\
 & temporary arrays, and Python object overhead \\
\bottomrule
\end{tabular}
}

\end{table}

The code and numerical records are available at \url{https://github.com/keiotakmin/endpoint-preserving-residual-correction}.

The ancillary comma-separated value (CSV) files packaged with this preprint provide losses before the aggregation in the main tables. The file \texttt{tableS01\_\allowbreak all\_\allowbreak conditions.csv} has one row per fixed-base condition, for 96 rows. Its \texttt{mse} and \texttt{mae} fields describe EPOC. The \texttt{static} and \texttt{elf} prefixes identify the uncorrected base and globally blended full ELF; \texttt{independent\_mse} is the coefficient-only correction. Percentage-field suffixes name the reference method, and \texttt{trace\_file} points to the archived block losses. The seven-arm file \texttt{table03\_\allowbreak all\_\allowbreak conditions.csv} has 672 rows, one for each method and fixed-base condition behind Table~\ref{tab:main_results}. Its archived \texttt{Proposed} method label denotes EPOC.

To reconstruct the Table~\ref{tab:main_results} overview from the seven-arm file, average \texttt{mse\_\allowbreak reduction\_\allowbreak pct} and \texttt{mae\_\allowbreak reduction\_\allowbreak pct} across the 96 rows of each method and take the median \texttt{state\_bytes}. For its dataset--backbone MSE and MAE panels, average the raw losses over the three seeds and two base-training phases within each dataset--backbone--method group. The \texttt{method} and \texttt{arm} fields identify the displayed comparator and its forecast variant.

The file \texttt{tableS02\_\allowbreak learned\_\allowbreak all.csv} has 864 rows: six series, two backbones, two warmups, three seeds, three adapter widths, and four forecast variants. Here \texttt{rank} is adapter width, \texttt{kind} is warmup, and \texttt{arm} identifies the uncorrected jointly trained \texttt{base}, raw \texttt{adapter}, globally blended \texttt{adapter\_global}, or globally blended EPOC (\texttt{boundary}). The \texttt{state\_bytes} field counts auxiliary correction arrays, so it is zero for the uncorrected base. The companion \texttt{tableS02\_\allowbreak width64\_\allowbreak training.csv} contains 288 width-64 rows, including the separately trained base-only runs used for the training comparison in Table~\ref{tab:learned_comparison}. In both files, the dataset code \texttt{bdg2} means BDG2 (Rat).

For the same-base percentages in Table~\ref{tab:learned_comparison}, average the three seed losses within each series--backbone--warmup--width--arm group in \texttt{tableS02\_\allowbreak learned\_\allowbreak all.csv}. Calculate $100(1-m_{\mathrm{boundary}}/m_{\mathrm{adapter\_global}})$ for each metric and each of the 12 series--backbone pairs, then average the 12 percentages. Table~\ref{tab:s_settings} collects the common evaluation and joint-training settings.

\FloatBarrier

\section{Setting Sensitivity}
\label{app:sensitivity}

Table~\ref{tab:s_sensitivity} lists the 13 one-factor settings summarized in Fig.~\ref{fig:overview_sensitivity}; the corresponding 1,248 condition-level rows are in \texttt{tableS05\_\allowbreak sensitivity\_\allowbreak conditions.csv}. Raising $K$ from 1 to 8 increases mean MSE reduction from 11.13\% to 16.40\% and median retained state from 2,056 to 12,080 B. With $K=4$, changing ridge strength, forgetting half-life, or blending window yields mean MSE reductions of 14.47--15.43\%. The shortest tested forgetting half-life, 32 blocks, gives 15.43\%; no forgetting gives 14.47\%.

\begin{table}[H]
\caption{One-factor sensitivity on 96 fixed-base conditions. Percentages average condition-wise reductions from Static with global blending; bytes are median retained arrays. Other settings match the reference.}
\label{tab:s_sensitivity}
\centering
\begingroup\fontsize{9}{11}\selectfont
\setlength{\tabcolsep}{4pt}
\renewcommand{\arraystretch}{1.12}
\begin{tabular}{@{}lrrr@{}}
\toprule
Setting & MSE (\%) & MAE (\%) & Bytes \\
\midrule
Reference ($K=4$) & 15.40 & 9.35 & 6,352 \\
$K=1$ & 11.13 & 6.47 & 2,056 \\
$K=2$ & 13.10 & 7.85 & 3,488 \\
$K=3$ & 14.49 & 8.79 & 4,920 \\
$K=6$ & 16.00 & 9.76 & 9,216 \\
$K=8$ & 16.40 & 10.08 & 12,080 \\
$\lambda=0.1$ & 15.19 & 9.17 & 6,352 \\
$\lambda=10$ & 15.18 & 9.46 & 6,352 \\
Half-life 32 & 15.43 & 9.72 & 6,352 \\
Half-life 512 & 14.82 & 8.96 & 6,352 \\
No forgetting & 14.47 & 8.76 & 6,352 \\
Window 8 & 15.24 & 9.35 & 5,968 \\
Window 128 & 14.98 & 9.07 & 7,888 \\
\bottomrule
\end{tabular}
\endgroup

\end{table}

\end{document}